\documentclass[letterpaper,12pt]{article}
\pdfoutput=1
\usepackage{amsmath}
\usepackage{amssymb}
\usepackage{algorithm,balance}
\usepackage{graphicx}
\usepackage{algorithm}
\usepackage{algorithmic}
\usepackage{amssymb,amsmath,latexsym,amsfonts,amsthm}
\usepackage{times}
\usepackage{xspace}
\usepackage{threeparttable} 
\usepackage{hyperref}

\newcommand*{\eg}{e.g.\@\xspace}
\newcommand*{\ie}{i.e.\@\xspace}

\title{CAN: Creative Adversarial Networks \\
Generating ``Art'' by Learning About Styles and Deviating from Style Norms\footnote{This paper is an extended version of a paper published on the eighth International Conference on Computational Creativity (ICCC), held in Atlanta, GA, June 20th-June 22nd, 2017.} 
}
\author{\href{mailto:elgammal@cs.rutgers.edu}{Ahmed Elgammal}$^1$\footnote{
Corresponding author: {Ahmed Elgammal \href{mailto:elgammal@cs.rutgers.edu}{elgammal@cs.rutgers.edu}}}  \;\; Bingchen Liu$^1$  \;\; Mohamed Elhoseiny$^2$  \;\; Marian Mazzone$^3$\\
\href{http://digihumanlab.rutgers.edu}{The Art \& AI Laboratory - Rutgers University} \\
$^1$ Department of Computer Science,
Rutgers University, NJ, USA\\
$^2$ Facebook AI Research, CA, USA \\
$^3$ Department of Art History,
College of Charleston, SC, USA }

\begin{document} 
\maketitle
\begin{abstract}
\begin{quote}

We propose a new system for generating art. The system generates art by looking at art and learning about style; and becomes creative by increasing the arousal potential of the generated art by deviating from the learned styles.  We build over Generative Adversarial Networks (GAN), which have shown the ability to learn to generate novel images simulating a given distribution. We argue that such networks are limited in their ability to generate creative products in their original design.  We propose modifications to its objective to make it capable of generating creative art by maximizing deviation from established styles and minimizing deviation from art distribution.  We conducted experiments to compare the response of human subjects to the generated art with their response to art created by artists. The results show that human subjects could not distinguish art generated by the proposed system from art generated by contemporary artists and shown in top art fairs.
\end{quote}
\end{abstract}

\section{Introduction}

Since the dawn of Artificial Intelligence, scientists have been exploring the machine's ability to generate human-level creative products such as poetry, stories, jokes, music, paintings, etc., as well as creative problem solving. This ability is fundamental to show that Artificial Intelligence algorithms are in fact intelligent. In terms of visual art, several systems have been proposed to automatically create art, not only in the domain of AI and computational creativity
(e.g.~\cite{baker1993evolving,dipaola2009incorporating,colton2015painting,heath2016before} ), but also in computer graphics (e.g. ~\cite{sims1991artificial}), and machine learning, (e.g. ~\cite{mordvintsev2015inceptionism,johnson2016perceptual}).

Within the computational creativity literature, different algorithms have been proposed
focused on investigating various and effective ways of exploring  the creative space. Several approaches have used an evolutionary process wherein the algorithm iterates by generating  candidates, evaluating them using a fitness function, and then modifying them to improve the fitness score for the next iteration (e.g.~\cite{machadoiterative,dipaola2009incorporating}).
Typically, this process is done within a genetic algorithm framework.  As pointed out by DiPaola and Gabora 2009, the challenge for any algorithm centers on ``how to write a logical fitness function that has an aesthetic sense''. 
Some earlier systems utilized a human in the loop with the role of guiding the process (e.g.~\cite{baker1993evolving,graf1995interactive}). In these interactive systems, the computer explores the creative space and the human plays the role of the observer whose feedback is essential in driving the process. Recent systems have emphasized the role of perception and cognition in the creative process~\cite{colton2008creativity,colton2015painting,heath2016before}.

 The goal of this paper is to investigate a computational creative system for art generation without involving a human artist in the creative process, but nevertheless involving human creative products in the learning process. An essential component in art-generating algorithms is relating their creative process to art that has been produced by human artists throughout time. We believe this is important because a human's creative process utilizes prior experience of and exposure to art. A human artist is continuously exposed to other artists' work, and has been exposed to a wide variety of art for all of his/her life. What remains largely unknown is how human artists integrate their knowledge of past art with their ability to generate new forms. A theory is needed to model how to integrate exposure to art with the creation of art.

Colin Martindale (1943-2008) proposed a psychology-based theory that explains new art  creation\cite{martindale1990clockwork}. He hypothesized that at any point in time, creative artists try to increase the arousal potential of their art to push against habituation. However, this increase has to be minimal to avoid negative reaction by the observers (principle of least effort). Martindale also hypothesized that style breaks happen as a way of increasing the arousal potential of art when artists exert other means within the roles of style. The approach proposed in this paper is inspired by Martindale's principle of least effort and his explanation of style breaks. Among  theories that try to explain progress in art, we find Martindale's theory to be computationally feasible. 

\begin{figure*}[thp]
\footnotesize
\center
\includegraphics[width=\textwidth]{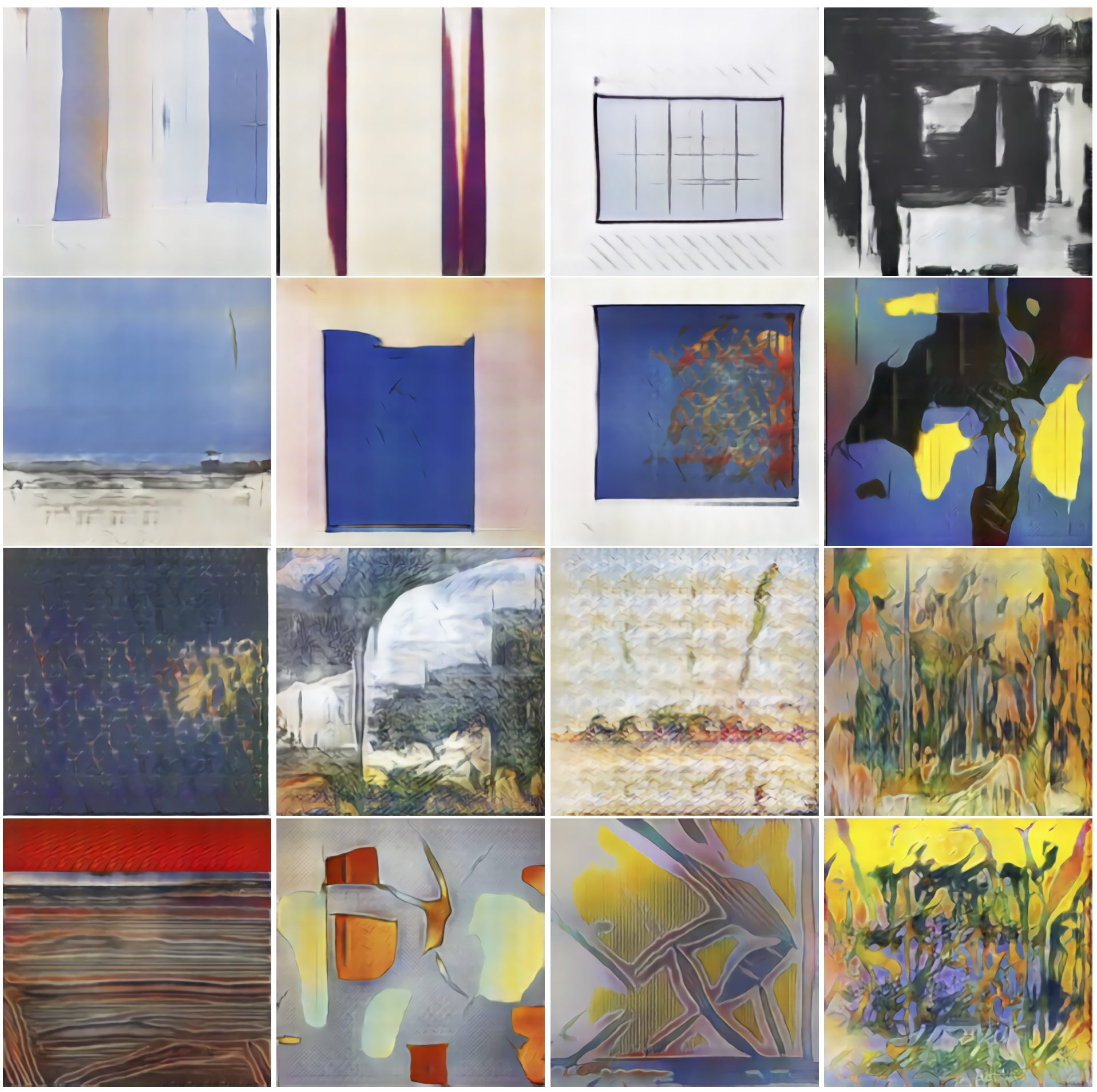} 
\label{fig:CAN1}
\caption{Example of images generated by CAN. The generated images vary from simple abstract ones to complex textures and compositions. More examples are shown in Figures~\ref{fig:CAN2} and \ref{fig:CANSamples}}.
\end{figure*}

Deep neural networks have recently played a transformative role in advancing artificial intelligence across various application domains. In particular, several generative deep networks  have been proposed that have the ability to generate novel images to emulate a given training distribution
Generative Adversarial Networks (GAN) have been quite successful in achieving this goal~\cite{goodfellow2014generative}.
We argue that such networks are limited in their ability to generate creative products in their original design. Inspired by Martindale's theory, in this paper we propose modifications to GAN's objective to make it able to generate creative art by maximizing deviation from established styles while minimizing deviation from art distribution.  Figure~\ref{fig:CAN1} shows sample of the generated images.

\section{Methodology}

\subsection{Background}
\label{S:Background}

The proposed approach is motivated from the theory suggested by D. E. Berlyne (1924-1976). Berlyne argued that the psychophysical concept of ``arousal'' has a great relevance for studying aesthetic phenomena~\cite{berlyne1971aesthetics}. ``Level of arousal'' measures how alert or excited a human being is. The level of arousal varies from the lowest level, when a person is asleep or relaxed, to the highest level when s/he is violent, in a fury, or in a passionate situation~\cite{berlyne1967arousal}. Among different
mechanisms of arousal, of particular importance and relevance to art are properties of external stimulus patterns~\cite{berlyne1971aesthetics}.

The term ``arousal potential'' refers to the properties of stimulus patterns that lead to raising arousal. Besides other psychophysical and ecological properties of stimulus patterns, Berlyne emphasized that the most significant arousal-raising properties for aesthetics are {\em novelty, surprisingness, complexity, ambiguity, and puzzlingness}. He coined the term {\em collative variables} to refer to these properties collectively.

Novelty refers to the degree a stimulus differs from what an observer has seen/experienced be- fore. Surprisingness refers to the degree a stimulus disagrees with expectation. Surprisingness is not necessarily correlated with novelty, for example it can stem from lack of novelty. Unlike novelty and surprisingness which rely on inter-stimulus comparisons of similarity and differences, complexity is an intra-stimulus property that increases as the number of independent elements in a stimulus grows. Ambiguity refers to the conflict between the semantic and syntactic information in a stimulus. Puzzlingness refers to the ambiguity due to multiple, potentially inconsistent, meanings.

Several studies have shown that people prefer stimulus with a moderate arousal potential~\cite{berlyne1967arousal,schneirla1959evolutionary}. Too little arousal potential is considered boring, and too much activates the aversion system, which results in negative response. This behavior is explained by the Wundt curve that correlates the arousal potential with the hedonic response~\cite{berlyne1971aesthetics,wundt1874grundzuge}.
Berlyne also studied arousal moderating mechanisms. Of particular importance in art is habituation, which refers to decreased arousal in response to repetitions of a stimulus~\cite{berlyne1971aesthetics}.

Martindale emphasized the importance of habituation in deriving the art-producing system~\cite{martindale1990clockwork}.  If artists keep producing similar works of arts, this directly reduces the arousal potential and hence the desirability of that art. Therefore, at any point of time, the art-producing system will try to increase the arousal potential of produced art. In other words, habituation forms a constant pressure to change art. However, this increase has to be within the minimum amount necessary to compensate for habituation without falling into the negative hedonic range, according to Wundt curve findings (``stimuli that are slightly rather than vastly supernormal are preferred''). Martindale called this the principle of ``least effort''. Therefore, there is an opposite pressure that leads to a graduated pace of change in art.

\subsection{Art Generating Agent}

We propose a model for an art-generating agent, and then propose a functioning model using a variant of GAN to make it creative. The agent's goal is to generate art with increased levels of arousal potential in a constrained way without activating the aversion system and falling into the negative hedonic range. In other words, the agent tries to generate art that is novel, but not too novel. This criterion is common in many computationally creative systems, however it is not easy to find a way to achieve that goal given the infinite possibilities in the creative space.

In our model the art-generating agent has a memory that encodes the art it has been exposed to, and can be continuously updated with the addition of new art. The agent uses  this encoded memory in an indirect way while generating new art with a restrained increase in arousal potential. While there are several ways to increase the arousal potential, in this paper we focus on building an agent that tries to increase the {\em stylistic ambiguity} and deviations from style norms, while at the same time, avoiding moving too far away from what is accepted as art. The agent tries to explore the creative space by deviating from the established style norms and thereby generates new art.

There are two types of ambiguities that are expected in the generated art by the proposed network; one is by design and the other one is inherent. Almost all computer-generated art might be ambiguous because the art generated typically does not have clear figures or an interpretable subject matter. Because of this, Heath et al argued that the creative machine would need to have perceptual ability (be able to see) in order to be able to generate plausible creative art~\cite{heath2016before}. This limited perceptual ability is what causes the inherent ambiguity. Typically, this type of ambiguity results in users being able to tell right away that the work is generated by a machine rather than a human artist. Even though several styles of art developed in the 20th century might lack recognizable figures or lucid subject matter, human observers usually are not fooled into confusing computer-generated art with human-generated art.

Because of this inherent ambiguity people always think of computer-generated art as being hallucination-like.  The Guardian commented on a the images generated by Google DeepDream~\cite{mordvintsev2015inceptionism} by ``Most, however, look like dorm-room mandalas, or the kind of digital psychedelia you might expect to find on the cover of a Terrence McKenna book''\footnote{Alex Rayner, the Guardian, March 28, 2016}. Others commented on it as being ``dazzling, druggy, and creepy''~\footnote{David Auerbach, Slate, July 23, 2015}. This negative reaction might be explained as a result of too much arousal, which results in negative hedonics according to the Wundt curve.

The other type of ambiguity in the art generated by the proposed agent is stylistic ambiguity, which is intentional by design.  The rational is that creative artists would eventually break from established styles and explore new ways of expression to increase the arousal potential of their art, as Martindale suggested. As suggested by DiPaola and Gabora, ``creators often work within a very structured domain, following rules that they eventually break free of''~\cite{dipaola2009incorporating}. 

The proposed art-generating agent is realized by a model called Creative Adversarial Network (CAN), which we will describe next. The network is designed to generate art that does not follow established art movements or styles, but instead tries to generate art that maximally confuses human viewers  as to which style it belongs to.

\subsection{GAN: Emulative and not Creative}

Generative Adversarial Network (GAN) has two sub networks, a generator and a discriminator. The discriminator has access to a set of images (training images). The discriminator tries to discriminate between ``real'' images (from the training set) and ``fake'' images generated by the generator.  The generator tries to generate images similar to the training set without seeing these images. The generator starts by generating random images and receives a signal from the discriminator whether the discriminator finds them real or fake. At equilibrium the discriminator should not be able to tell the difference between the images generated by the generator and the actual images in the training set, hence the generator succeeds in generating images that come from the same distribution as the training set.  

Let us now assume that we trained a GAN model on images of paintings. Since the generator is trained to generate images that fool the discriminator to believe it is coming from the training distribution, ultimately the generator will just generate images that look like already existing art. There is no motivation to generate anything creative. There is no force that pushes the generator to explore the creative space. Let us think about a generator that can cheat and already has access to samples from the training data. In that case the discriminator will right away be fooled into believing that the generator is generating art, while in fact it is already existing art, and hence not novel and not creative.

There have been extensions to GANs that facilitate generating images conditioned on  categories (e.g.,~\cite{DCGAN16}) or captions (e.g.,~\cite{reed2016generative}). We can think of a GAN that can be designed and trained to generate images of different art styles or different art genres by providing such labels with training. This might be able to generate art that looks like, for example, Renaissance, Impressionism, or Cubism. However that does not lead to anything creative either.  No creative artist will create art today that tries to emulate the Baroque or Impressionist style, or any traditional style, unless doing so ironically. According to Berlyne and Martindale, artists would try to increase the arousal potential of their art by creating novel, surprising, ambiguous, and/or puzzling art. This highlights the fundamental limitation of using GANs in generating creative works.

\subsection{From being Emulative to being Creative}

In the proposed Creative Adversarial Network (CAN), the generator is designed to receive two signals from the discriminator that act as two contradictory forces to achieve three points: 1) generate novel works, 2) the novel work should not too novel, i.e., it should not be too far away from   the distribution or it will generate too much arousal, thereby activating the aversion system and falling into the negative hedonic range according to the Wundt curve, 3) the generated work should increase the stylistic ambiguity.

Similar to Generative Adversarial Networks (GAN), the proposed network has two adversary networks, a discriminator and a generator. The discriminator has access to a large set of art associated with style labels (Renaissance, Baroque, Impressionism, Expressionism, etc.) and uses it to learn to discriminate between styles. The generator does not have access to any art. It generates art starting from a random input, but unlike GAN, it receives two signals from the discriminator for any work it generates. The first signal is the discriminator's classification of ``art or not art''. In traditional GAN, this signal enables the generator to change its weights to generate images that more frequently will deceive the discriminator as to whether it is coming from the same distribution. Since the discriminator in our case is trained on art, this will signal whether the discriminator thinks the generated art is coming from the same distribution as the actual art it knows about. In that sense, this signal flags whether the discriminator thinks the image presented to it is ``art or not art''. Since the generator only receives this signal, it will eventually converge to generate images that will emulate art.

\begin{figure*}[t]
\center
  \includegraphics[width=6in]{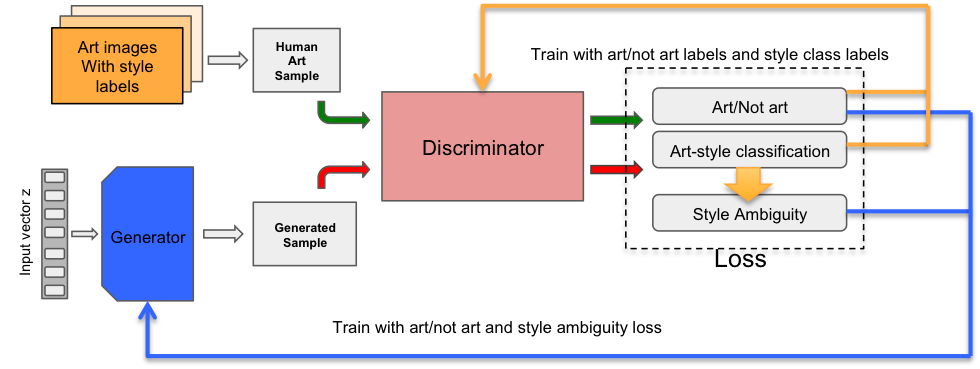}
  \caption{\footnotesize Block diagram of the CAN system.  
}
  \label{fig:BD}
\end{figure*}

The second signal the generator receives is a signal about how well the discriminator can classify the generated art into established styles. If the generator generates images that the discriminator thinks  are art and also can easily classify into one of the established styles, then the generator would have fooled the discriminator into believing it generated actual art that fits within established styles. In contrast, the creative generator will try to generate art that confuses the discriminator. On one hand it tries to fool the discriminator to think it is ``art,'' and on the other hand it tries to confuse the discriminator about the style of the work generated.

These two signals are contradictory forces, because the first signal pushes the generator to generate works that the discriminator accepts as ``art,'' however if it succeeds within the rules of established styles, the discriminator will also be able to classify its style. Then the second signal will heftily penalize the generator for doing that.   This is because   the second signal pushes the generator to generate style-ambiguous works. Therefore, these two signals together should push the generator to explore parts of the creative space that lay close to the distribution of art (to maximize the first objective), and at the same time maximizes the ambiguity of the generated art with respect to how it fits in the realm of standard art styles.


\section{Technical Details}

 \subsection{Generative Adversarial Networks}
 Generative Adversarial Network (GAN)~\cite{goodfellow2014generative} is one of the most successful image synthesis models in the past few years. GAN is typically trained by setting a game between two players. The first player, called the generator, $G$, generates samples that are intended to come from the same probability distribution as the training data (\ie $p_{data}$), without having access to such data. The other player, denoted as the  discriminator,  $D$,  examines the samples to determine whether they are coming from $p_{data}$ (real) or not (fake). Both the discriminator and the generator are typically modeled as deep neural networks.  The training procedure is similar to a two-player min-max game with the following objective function
\begin{equation}
\small
\label{eq:GAN_org}
\begin{aligned}
\min_{G} \max_{D} V(D,G) = \; & \mathbb{E}_{x \sim {p_{data}}} [\log D(x)] \; + \\
& \mathbb{E}_{z \sim {p_{z}}} [\log(1 -  D(G(z)))],
\end{aligned}
\end{equation}
where  $z$ is a noise vector sampled from distribution $p_{z}$ (\eg, uniform or Gaussian distribution) and $x$ is a real image from the data distribution $p_{data}$. In practice, the discriminator and the generator are alternatively optimized for every batch. The discriminator aims at maximizing  Eq~\ref{eq:GAN_org} by minimizing  $-\mathbb{E}_{x \sim {p_{data}}} [\log D(x)] -  \mathbb{E}_{z \sim {p_{z}}} [\log(1 - D(G(z)))$, which improves the utility of the $D$ as a fake vs. real image detector. Meanwhile, the generator  aims at minimizing  Eq~\ref{eq:GAN_org}  by maximizing $log(D(G(z))$, which works better than  $-log(1-D(G(z))$ since it provides stronger gradients. By optimizing $D$ and $G$ alternatively, GAN is trained to generate images that emulate the training distribution. 

\begin{figure*}[t]
\center
  \includegraphics[width=2.5in]{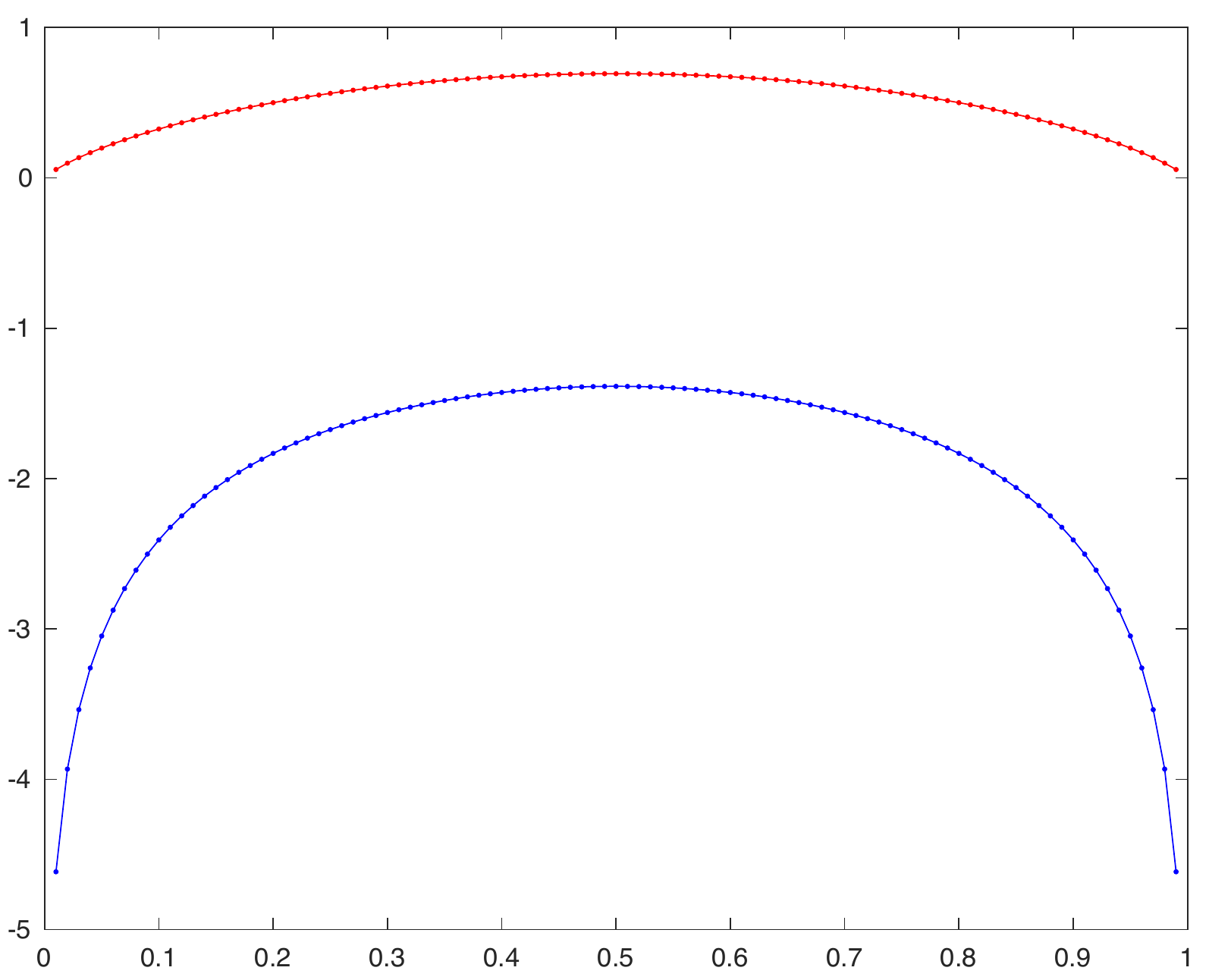}
  \caption{\footnotesize Style ambiguity cross entropy loss function (two class case). Red curve: entropy, Blue curve: cross entropy with a uniform distribution (inverted here for illustration). Both functions are maximized when the classes are equiprobable. In contrast to entropy, which goes to zero at the boundaries, the inverted cross entropy goes to negative infinity at the boundary, which causes hefty penalty for samples classified correctly.}
  \label{fig:crossentropyfn}
\end{figure*}

 \subsection{Creative Adversarial Networks}
 
 We modified the GAN loss function to achieve the vision explained in the previous section. Figure~\ref{fig:BD} illustrates the architecture. We added a style classification loss and a style ambiguity loss. Maximizing the stylistic ambiguity can be achieved by maximizing the style class posterior entropy.  Hence, we need to design the loss such that the generator $G$ produces an image $x\sim {p_{data}}$ and, meanwhile, maximizes the entropy of $p(c|x)$ (\ie style class posterior) for the generated images.
 However, instead of maximizing the class posterior entropy, we minimize the cross entropy between the class posterior and a uniform target distribution. Similar to entropy that is maximized when the class posteriors (i.e., $p(c|G(z))$) are equiprobable, cross entropy with uniform target distribution will be minimized when the classes are equiprobable.  So both objectives will be optimal when the classes are equiprobable. However, the difference is that the cross entropy will go up sharply at the boundary since it goes to infinity if any class posterior approaches 1 (or zero), while entropy goes to zero at this boundary condition (see Figure~\ref{fig:crossentropyfn}). Therefore, using the cross entropy results in a hefty penalty if the generated image is classified to one of the classes with high probability. This in turn would generate very large loss, and hence large gradients if the generated images start to be classified to any of the style classes with high confidence. Hence, we can redefine the cost function with a different adversarial objective as
\begin{equation}
\small
\begin{split}
\min_{G}& \max_{D} V(D,G) = \;  \\ 
&\mathbb{E}_{x,\hat{c} \sim {p_{data}}} [\log D_r(x) + \log D_c(c=\hat{c}|x)] \; + \\
& \mathbb{E}_{z \sim {p_{z}}} [\log(1 -  D_r(G(z))) - \sum_{k=1}^K  \big( \frac{1}{K} log(D_c(c_k|G(z)) +  \\
& \,\,\,\,\,\,\,\,\,\,\,\,\,\,\,\,\,\,\,\,\,\,\,\,\,\,\,\,\,\,\,\,\,\,\,\,\,\,\,\,\,\,\,\,\,\,\,\,\,\,\,\,\,\,\,\,\, (1- \frac{1}{K}) log(1- D_c(c_k|G(z)) \big)],
\label{eqcostcan2}
\end{split}
\end{equation}
where  $z$ is a noise vector sampled from distribution $p_{z}$ (\eg, uniform or Gaussian distribution) and $x$ and $\hat{c}$ are a real image  and its corresponding style label from the data distribution $p_{data}$.  $D_r(\cdot)$ is the transformation function that tries to discriminate between real art and generated images. $D_c(\cdot)$ is the the function that discriminates between different style categories and estimates the style class posteriors (i.e., $D_c(c_k| \cdot) = p(c_k| \cdot)$).

\noindent \textbf{Discriminator Training:} In Eq~\ref{eqcostcan2}, the discriminator $D$ encourages  maximizing Eq~\ref{eqcostcan2} by minimizing  $-\mathbb{E}_{x \sim {p_{data}}} [\log D_r(x) + \log D_c(c=\hat{c}|x)]$ for the real images and  $-\mathbb{E}_{z \sim {p_{z}}} [\log(1 -  D_r(G(z))) ]$ for the generated images.  The discriminator is trained to, not only discriminate the real art samples from the generated (fake) ones, but also to identify their style class though the K-way loss (where K is the number of style classes). Therefore, the discriminator is simultaneously learning about both the art distribution and art styles. 

\noindent \textbf{Generator Training:} The generator $G$ encourages minimizing  Eq~\ref{eqcostcan2}  by maximizing $log(1-D_r(G(z)) - \sum_{k=1}^K ( \frac{1}{K} log(D_c(c_k|G(z)) + (1- \frac{1}{K}) log(1- D_c(c_k|G(z)) $. This pushes the generated images to look as real art (first term) and meanwhile to have a large cross entropy for $p(c|G(z))$ with a uniform distribution to maximize style ambiguity (second term).  Note that the CAN generator does not require any class labels, similar to unconditional generative model. 


\noindent \textbf{Model Architecture:} 
Algorithm~\ref{alg_label} illustrates CAN training process. 
The Generator $G$ and similar to DCGAN architecture~\cite{DCGAN16}, first  $z \in \mathbb{R}^{100}$ normally sampled from 0 to 1 is up-sampled to a $4 \times $ spatial extent convolutional representation with 2048 feature maps resulting in a  $4\times4\times 2048$ tensor. Then a series of four fractionally-stride convolutions (in some papers, wrongly called deconvolutions). Finally, convert this high level representation into a $256 \times 256$ pixel image. In other words, starting from $z \in \mathbb{R}^{100} \to 4 \times 4 \times 1024 \to 8\times 8\times 1024 \to 16 \times 16 \times 512 \to  32 \times 32 \times 256 \to  64 \times 64 \times 128 \to  128 \times 128 \times 64 \to  256 \times 256 \times 3$ (the generated image size). As described earlier, the discriminator has two losses (real/fake loss and multi-label loss). The discriminator in our work starts by a common body of convolution layers followed by two heads (one for the real/fake loss and one for the multi-label loss). \textit{The common body} of convolution layers is composed of a series of six convolution layers (all with stride 2 and 1 pixel padding). conv1 (32 $4\times 4$  filters), conv2 (64 $4\times 4$  filters, conv3 (128 $4\times 4$  filters,  conv4 (256 $4\times 4$  filters, conv5 (512 $4\times 4$  filters, conv6 (512 $4\times 4$  filters). Each convolutional layer is followed by a leaky rectified activation (LeakyRelU)~\cite{maas2013rectifier,xu2015empirical} in all the layers of the discriminator. After passing a image to the common conv $D$ body, it will produce a feature map or size ($4\times 4 \times 512$).  \textit{The real/fake $D_r$ head} collapses the ($4\times 4 \times 512$) by a fully connected to produce $D_r(c|x)$ (probability of image coming for the real image distribution). The \textit{multi-label probabilities  $D_c (c_k|x)$ head} is produced by passing the($4\times 4 \times 512$) into 3 fully collected layers sizes $1024$, $512$, $K$, respectively, where $K$ is the number of style classes.

\noindent \textbf{Initialization and Training:} The weights were initialized from a zero-centered Normal distribution with standard deviation 0.02. We used a mini-batch size of 128 and used mini-batch stochastic gradient descent (SGD) for training  with 0.0001 as learning rate.  In the LeakyReLU, the slope of the leak was set to 0.2 in all models. While previous GAN work has used momentum to accelerate training, we used the Adam optimizer and trained the model for 100 epochs (100 passes over the training data). To stabilize the training, we used Batch Normalization~\cite{ioffe2015batch} that normalizing the input to each unit to have zero mean and unit variance. We performed data augmentation by adding 5 crops within for each image (bottom-left, bottom-right, mid, top-left, top-right) on our image dataset. The width and hight of each crop is 90\% of the width and the hight of the original painting. 

\begin{algorithm}[t]
	\begin{algorithmic}[1]
		\STATE {\bfseries Input:} mini-batch images $x$, matching label $\hat{c}$, number of training batch steps $S$
		\FOR{$n=1$ {\bfseries to} $S$}
		\STATE $z \sim \mathcal{N}(0,1)^{Z}$ \COMMENT{Draw sample of random noise}
		\STATE $\hat{x} \gets G(z)$ \COMMENT{Forward through generator}
		\STATE $s_{D}^{r} \gets D_r(x)$ \COMMENT{real image, real/fake loss }
		\STATE $s_{D}^{c} \gets D_c(\hat{c}|x)$ \COMMENT{real image, multi class loss}
		\STATE $s_{G}^{f} \gets D_r(\hat{x})$ \COMMENT{fake image, real/fake loss}
        \STATE $s_{G}^{c} \gets \sum_{k=1}^K \frac{1}{K} log(p(c_k|\hat{x}) + (1- \frac{1}{K} ) (log(p(c_k|\hat{x}))$ \COMMENT{fake image  Entropy loss}
		\STATE $\mathcal{L}_{D} \gets \log(s_{D}^{r}) + \log(s_{D}^{c}) + \log(1-s_{G}^{f})$
		\STATE $D \gets D - \alpha \partial{\mathcal{L}}_{D} / \partial D$ \COMMENT{Update discriminator}
		\STATE $\mathcal{L}_G \gets \log(s_{G}^{f}) - s_{G}^{c}$
		\STATE $G \gets G - \alpha \partial{\mathcal{L}}_G / \partial G$ \COMMENT{Update generator}
		\ENDFOR
	\end{algorithmic}
	\caption{CAN training algorithm with step size $\alpha$, using mini-batch SGD for simplicity.\label{alg_label}}
\end{algorithm}

\section{Results and Validation}

\subsection{Training the model}

We trained the networks using paintings from the publicly available WikiArt dataset~\footnote{https://www.wikiart.org/}. This collection (as downloaded in 2015) has images of 81,449 paintings from 1,119 artists ranging from the Fifteenth century to Twentieth century. Table~\ref{style-number-table} shows the number of images used in training the model from each style. 

\begin{table}
\centering
\caption{ Artistic Styles Used in Training } 
\label{style-number-table}
\begin{tabular}{|l|r|l|r|}
\hline
Style name                 & Image number & Style name                 & Image number \\\hline
Abstract-Expressionism     	&	2782	&	Mannerism-Late-Renaissance 	&	1279 \\\hline
Action-Painting            	&	98	&	Minimalism                 	&	1337 \\\hline
Analytical-Cubism          	&	110	&	Naive Art-Primitivism      	&	2405 \\\hline
Art-Nouveau-Modern         	&	4334	&	New-Realism                	&	314 \\\hline
Baroque                    	&	4241	&	Northern-Renaissance       	&	2552 \\\hline
Color-Field-Painting       	&	1615	&	Pointillism                	&	513 \\\hline
Contemporary-Realism       	&	481	&	Pop-Art                    	&	1483 \\\hline
Cubism                     	&	2236	&	Post-Impressionism         	&	6452 \\\hline
Early-Renaissance          	&	1391	&	Realism                    	&	10733 \\\hline
Expressionism              	&	6736	&	Rococo                     	&	2089 \\\hline
Fauvism                    	&	934	&	Romanticism                	&	7019 \\\hline
High-Renaissance           	&	1343	&	Synthetic-Cubism           	&	216 \\\hline
Impressionism              	&	13060	&	Total	&	75753 \\\hline
\end{tabular}

\end{table}

\subsection{Qualitative Validation}
\label{S:QualExp}

\begin{figure*}[thp]
\footnotesize
\center
\includegraphics[width=0.9\textwidth]{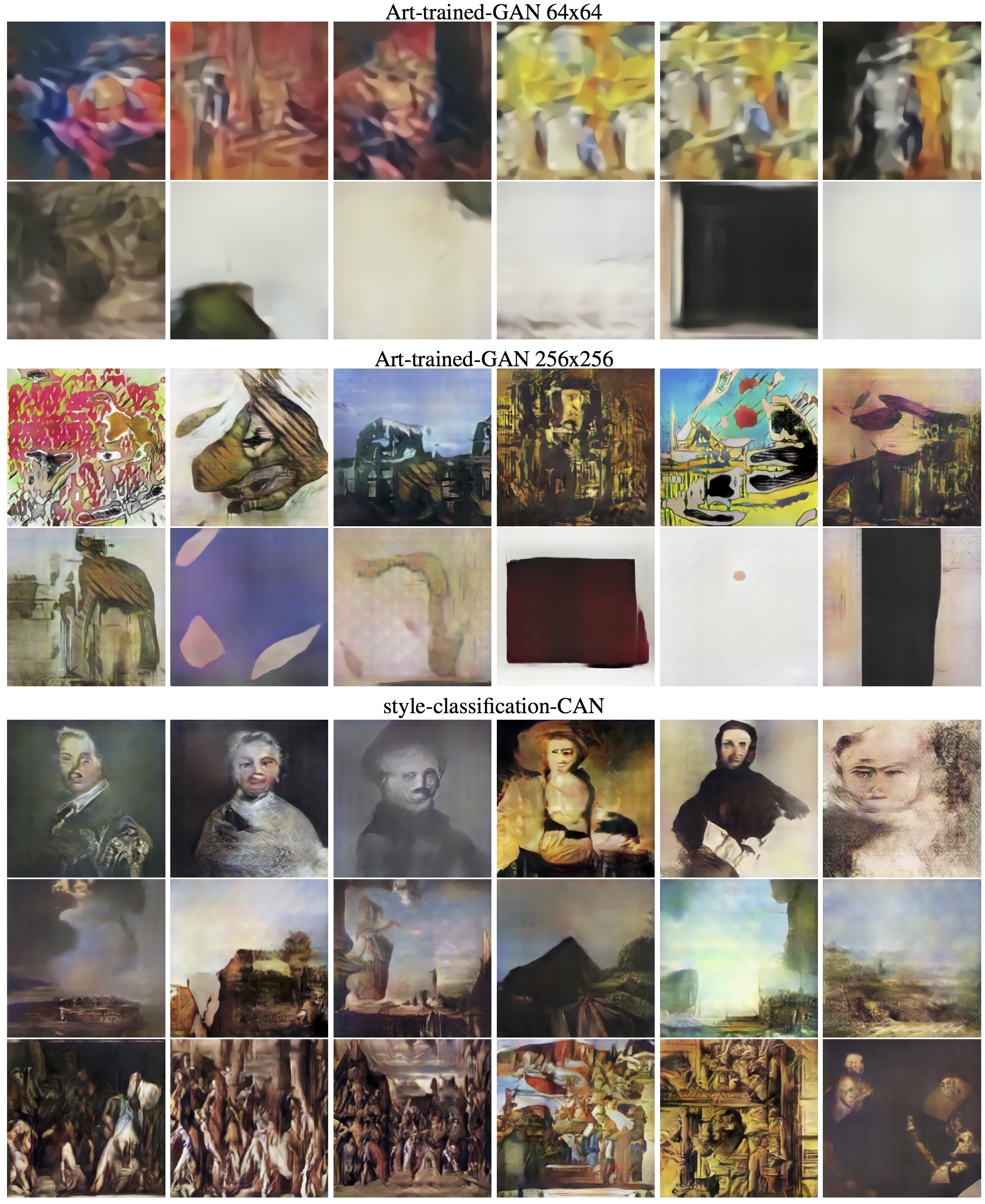} 
\caption{Images from three baselines. Top Panel: Sample images generated by art-trained DCGAN 64x64 baseline model (top 6 and bottom 6 ranked images are shown according to human subject likeness rating in Experiment I). Middle Panel: Sample images generated by art-trained DCGAN 256x256 baseline model (top 6 and bottom 6 ranked images are shown according to human subject likeness rating in Experiment II). Bottom Panel: Sample images generated by style-classification CAN model, a variant of CAN with style classification loss and without the style ambiguity loss. The images show recognizable genres such as portraits, landscapes, etc. }
\label{fig:DifferentModels}
\end{figure*}

\begin{figure*}[thp]
\footnotesize
\center
\includegraphics[width=\textwidth]{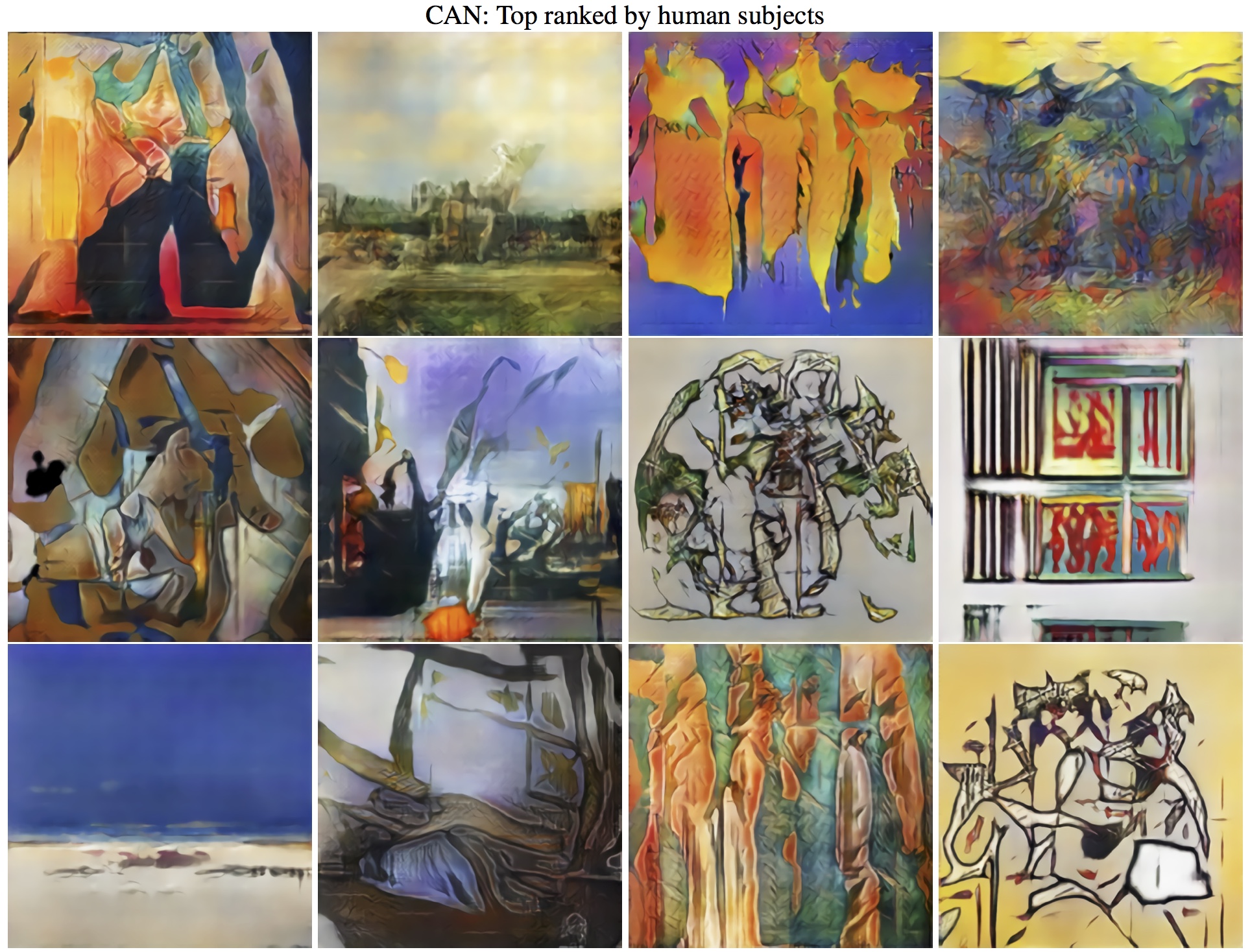} 
\includegraphics[width=\textwidth]{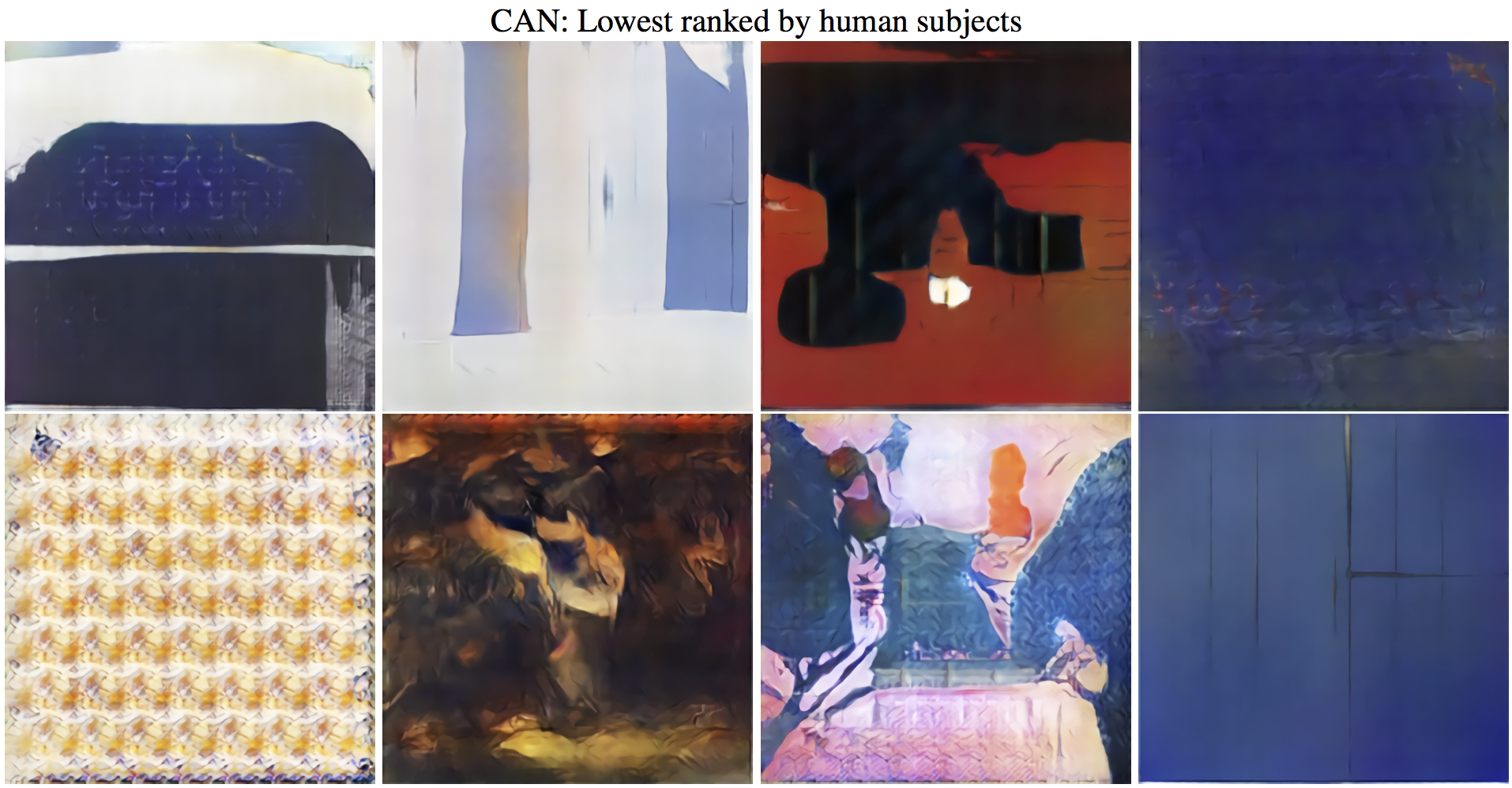} 
\label{fig:CAN2}
\caption{Example of images generated by CAN. Top:  Images ranked high in ``likeness'' according to human subjects. Bottom: Images ranked the lowest by human subjects.
}
\end{figure*}

Assessing the creativity of artifacts generated by the machine is an open and hard question. As noted by Colton 2008, aesthetic assessment of an artifact is different from the creativity assessment~\cite{colton2008creativity}.  
Figures~\ref{fig:CAN1}, Figures~\ref{fig:CAN2} and \ref{fig:CANSamples} show samples of the generated images by the proposed CAN model. The images generated by CAN do not look likes traditional art, in terms of standard genres (portrait, landscapes, religious paintings, still life,  etc.). We also do not see any recognizable figures. Many of the images seems abstract.  Is that simply because it fails to emulate the art distribution, or is it because it tries to generate novel images? Is it at all creative?  These are hard questions which we will try to get an insight to help answering them. 

We evaluated the images generated by three baseline models in comparison to the ones generated by the proposed CAN model to gain an insight into the differences qualitatively and quantitatively. All the models were trained on the same art dataset, as shown in table~\ref{style-number-table}.

The first two baselines are variants of the original DCGAN~\cite{DCGAN16} model trained on art data.
Since GAN aims to emulate the training distribution, we should expect training the model on art data would result in generating images that show recognizable figures, subject matters, art genres, and styles. The first baseline model is the original DCGAN~\cite{DCGAN16} model trained on art data. This model generates images with 64x64 resolution. Although trained on art data, this model failed to generate images that emulate the trained art. The generated samples  did not show any recognizable figures or art genres or styles. The top panel of Figure~\ref{fig:DifferentModels} shows samples of the generated images; here we show the six top and six bottom  ranked images based on human subject evaluation in Experiment I below. 

The second baseline model is the original DCGAN~\cite{DCGAN16} model after adding two more layers to the generator to increase the resolution to 256x256, i.e., the generator here has the same exact architecture as the CAN model. We also trained this model on the art collection. The generated samples show significant improvement; we can clearly see aesthetically appealing compositional structures and color contrasts in the resulting images. However, the generated images also did not show any recognizable figures, subject matters or art genres. The middle panel of Figure~\ref{fig:DifferentModels} shows samples of the generated images; here we show the six top and six bottom  ranked images based on human subject evaluation in Experiment II below.

The third model is a variant of the proposed CAN model with the style classification loss (without the style ambiguity loss). In that model the discriminator learns to discriminate between style classes along learning the art distribution. The generator has exactly the same loss as the GAN model, i.e., only try to deceive the discriminator by emulating the training distribution. In other words, unlike the two first baselines that learn about art (art/nor art), this model also learn about styles classes in art. We refer to this model by style-classification-CAN. The generated images of that model show significant improvement in actually emulating the art distribution, in the sense that we can see lots of hallucination of portraits, landscapes, architectures, religious subject matter, etc. We didn't see any of that on the first two baselines. The bottom panel of Figure~\ref{fig:DifferentModels} shows samples of the generated images, which show portraits, landscapes, architecture elements, religious themes, etc. This baseline shows that the CAN model, without the style ambiguity, can better emulate the art distribution by learning about style classes, however not creative (Experiment IV below is designed to quantitatively validate this claim).

Figures~\ref{fig:CAN2} shows samples of images generated by CAN. The figure shows top ranked and lowest ranked images according to human subjects. 
In contrast to the aforementioned three baselines, the proposed CAN model generates images that 
can be characterized as novel and not emulating the art distribution, however, aesthetically appealing.   Although the generated images of the CAN model do not show typical figures, genres, styles, or subject matter (similar to the first two baselines anyway), we cannot say that this is because it cannot emulate the art distribution, since, simply, removing the style ambiguity loss reduces the model to the third baseline (style-classification-CAN) which is successful in generating such elements.   So we can claim that the style ambiguity loss forces the network to try to generate novel images and, in the same time, stay closer to the art distribution (by being aesthetically appealing). We will try to test this claim quantitatively by a series of human subject experiments in the next section. 

\subsection{Quantitative Validation}



\begin{figure*}[tph]
\includegraphics[width=\textwidth]{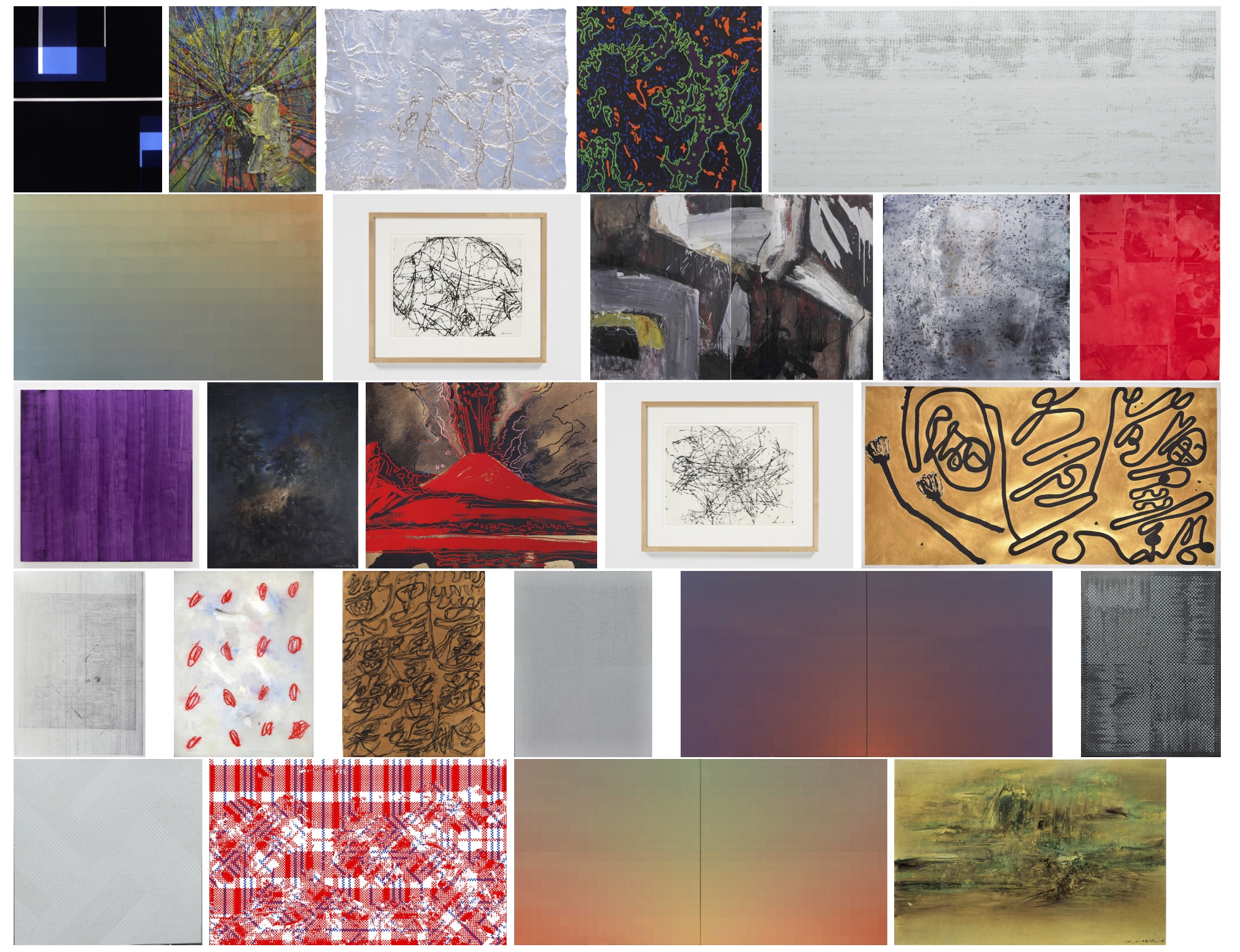} 
\caption{Art Basel Set: a collection of 25 paintings selected from Art Basel 2017 art fair. Artist and year in order:
 Richard Caldicott 	2003,
Jigger Cruz 	2016,
Leonardo Drew 	2015,
Cenk Akaltun 	2015,
Lang Li 	2014,
Xuerui Zhang	2015,
David Smith 	1956,
 Kelu Ma 	1989,
Xie Nanxing 	2013,
Panos Tsagaris 	2015,
Heimo Zobernig 	2014,
Zao Wou-Ki 	1958,
Andy Warhol 	1985,
David Smith 	1956,
Wei Ligang 	2014,
KONG Chun Hei 	2016,
Ye Yongqing 	2015,
 Wei Ligang 	2010,
Xiaorong Pan 	2015,
Xuerui Zhang 	2016,
Xiaorong Pan 	2015,
Xiaorong Pan 	2015,
Xu Zhenbang	2015,
 Xuerui Zhang 	2016,
Zao Wou-Ki 	1963.
}
\label{fig:AB}
\end{figure*}

We conducted human subject experiments to evaluate aspects of the creativity of the proposed model. 
The goal of these experiments is to test whether human subjects would be able to distinguish whether the art is generated by a human artist or by a computer system, as well as to rate aspects of that art.
However, the hard question is which art by human artists we should use for this comparison. Since the goal of this study is to evaluate the creativity of the artifacts produced by the proposed system, we need to compare human response to such artifacts with art that is considered to be novel and creative at this point in time. If we compare the produced artifacts to, for example, Impressionist art, we would be testing the ability of the system to emulate such art, and not the creativity of the system. Therefore we collected two sets of works by real artists, as well as four machine-generated sets as follows:

\begin{enumerate}
\item Abstract Expressionist Set: A collection of 25 paintings by Abstract Expressionist masters made between 1945-2007, many of them by famous artists. This set was previously used in recent studies to compare human and machine's ability to distinguish between abstract art created by artists, children or animals~\cite{snapper2015your,shamir2016distinguishing}. We use this set as a baseline set. Human subjects are expected to easily determine that these are created by artists based on familiarity. We used Abstract Expressionist art in particular because they lack recognizable figures or lucid subject matter. Existence of figures or clear subject matter might directly bias the subjects to conclude that such paintings are done by human when contrasted to the generated images which lacks such figures.  

\item Art Basel 2016 Set: This set consists of 25 paintings of various artists that were shown in Art Basel 2016, which is the flagship art fair for contemporary art world wide. Being shown in Art Basel 2016 is an indication that these are art works at the frontiers of human creativity in paintings, at least as judged by the art experts and the art market. We selected this set at random after excluding art that has clear figures or obvious brush strokes which might bias the subjects. The collection is shown in Figure~\ref{fig:AB}.

\item DCGAN Sets: We used two sets of images generated by the state-of-the art Deep Convolution GAN (DCGAN) architecture~\cite{DCGAN16}, as described in the baseline models in Section~\ref{S:QualExp}. The first set contains 100 images generated at 64x64 resolution. The second set consists of 76 images generated at 256x256 resolution.   

\item Style Classification CAN  Set ({\it sc}-CAN): a set of 100 images generated by a variant of the model that involves style classifier at the discriminator, as explained in Section~\ref{S:QualExp}. We mainly used this set to compare the effectiveness of adding style ambiguity to the proposed loss function. 

\item  CAN Set: a set of 125 images generated by the proposed model.

\end{enumerate}
We used the same set of training images for all the models and we conducted three human subject experiments as follows. All generated images were upscaled to 512x512 resolution using a super-resolution algorithm. Artist images were also resized to 512x512 resolution.

\begin{table}
\centering
\caption{ Means and standard deviations of responses of Experiment I } 
\label{exp1-result}
\begin{tabular}{|c|c|c|}
\hline
Painting set & Q1 (std) & Q2 (std)    \\\hline
CAN & 53\% (18\%)$^\dag$ & 3.2 (1.5) $^\ddag$     \\\hline
DCGAN~\cite{DCGAN16} (64x64)& 35\% (15\%) $^\dag$ & 2.8 (0.54) $^\ddag$    \\\hline
Abstract Expressionist   & 85\% (16\%) & 3.3 (0.43)      \\\hline
Art Basel 2016               & 41\% (29\%) & 2.8 (0.68)     \\\hline
Artist sets combined    & 62\% (32\%) & 3.1 (0.63)     \\\hline
\multicolumn{3}{|l|}{\footnotesize All images are resized to 512x512 resolution} \\ 
\multicolumn{3}{|l|}{\footnotesize $^\dag$ Q1{\it t}-test (CAN vs. DCGAN) p-value = $1.9932e-15$ }\\
\multicolumn{3}{|l|}{\footnotesize $^\ddag$ Q2 {\it t}-test (CAN vs. DCGAN) p-value = $9.3634e-06$} \\
\hline
\end{tabular}
\end{table}

\subsection*{Experiment I:} 
The goal of this experiment is to test the ability of the system to generate art that human users could not distinguish from top creative art that is being generated by artists today. We used four image sets (Abstract Expressionist, Art Basel, CAN and DCGAN(64x64)). In this experiment each subject was shown one image at time selected from the four sets of images and asked:
\begin{description}
\item[Q1:] Do you think the work is created by an artist or generated by a computer? The user has to choose one of two answers: artist or computer.
\item[Q2:] The user asked to rate how they like the image in a scale 1 (extremely dislike) to 5 (extremely like).
\end{description}

\begin{figure}
\center
  \includegraphics[width=0.5\linewidth]{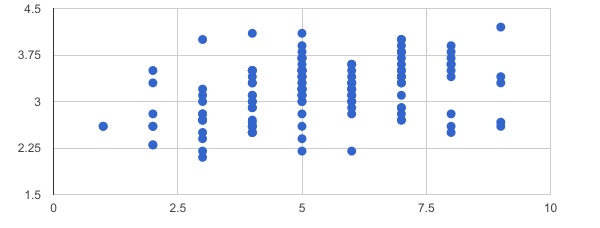}
  \caption{Experiment I (Q1 vs. Q2 responses)}
  \label{fig:scatter-plot}
\end{figure}

Results: 18 MTurk users participated in this experiment, where for each image we received 10 distinct responses. The results are summarized in Table~\ref{exp1-result}.  There are several conclusions we can draw from these results: 1) As expected, subjects rated the Abstract Expressionist set higher as being created by an artist (85\%).
 2) The proposed CAN model out-performed the  GAN model in generating images that human subjects think are generated by artist (53\% vs. 35\%). Human subjects also liked the images generated by CAN more than GAN (3.2 vs. 2.8) 
  We performed two-sample {\it t}-test to determine the statistical significance of these results, with the null hypothesis being that the subjects' responses for both CAN and GAN are coming from the same distribution. {\it t}-test rejected this hypothesis with p-value = $1.9932e-15$ and $9.3634e-06$ for questions 1 and 2 respectively. This test highlights that the difference is statistically significant.  
 Of course we cannot obviously conclude from this results that CAN is more creative than GAN. A  system that would perfectly copy human art, without being innovative, would score higher in that question. However, we can exclude this possibility since the generated images by both CAN and GAN are not copying human art as was explained in Section~\ref{S:QualExp}. 
 3) More interestingly, human subject rated the images generated by CAN higher as being created by a human than the ones from the Art Basel set (53\% vs. 41\%) when combining the two sets of art created by artists, the images generated by CAN scored only 9\% less (53\% vs. 62\%). Figure~\ref{fig:CANSamples} shows the top ranked CAN images according to subjects' responses.  Figure~\ref{fig:scatter-plot} shows a scatter plot of the responses for the two questions, which interestingly shows weak correlation between the likeness rating and whether subjects think it is by an artist or a computer.

\subsection{Experiment II:}
To confirm the results of experiment I we designed another experiment where in each survey we showed an image and asked subjects a series of questions first before asking the question whether the shown image is generated by a human artist or computer. We hypothesized that if that question is asked first, this would have a higher chance of subjects answering it randomly, and deferring it till after a series of other questions about the image would lead to a more constructive response. 

The questions are based on the collative variables as explained in Section~\ref{S:Background} and specified as:
\begin{description}
\item [Q1] How do you like this image: 1-extremely dislike, 2-dislike, 3-Neutral, 4-like, 5-extremely like.
\item [Q2] Rate the novelty of the image: 1-extremely not novel, 2-some how not novel, 3-neutral, 4- somehow novel, 5-extremely novel.
\item [Q3] Do you find the image surprising: 1-extremely not surprising, 2-some how not surprising, 3-neutral, 4-some how surprising, 5-extremely surprising.
\item [Q4] Rate the ambiguity of the image. I find this image: 1-extremely not ambiguous, 2-some how not ambiguous, 3-neutral, 4-some how ambiguous, 5-extremely ambiguous.
\item [Q5] Rate the complexity of the image. I find this image: 1-extremely simple, 2-some how simple, 3-neutral, 4-somehow complex, 5-extremely complex
\item [Q6] Do you think the image is created by an artist or generated by computer?
\end{description}

We used the same sets of images as in experiment I except that we changed the GAN set to a new set generated by DCGAN model with output resolution 256x256, which is the same resolution as the CAN model. Both the CAN and GAN sets were upsampled to 512x512.  All artists sets are resized to 512x512 as well. 

Results: We conducted this experiment using Amazon MTurk, where for each image we received 10 distinct responses. Table~\ref{exp2-result} summarize the results. In general there is no much differences in the ratings for the different sets for questions 1 to 5. However the responses to Q6 are significantly different for each dataset. 
The results show that CAN out-perform GAN in generating images that human subjects think are generated by artist (75\% vs. 65\%). This difference is statistically significant ({\it t}-test p-value=$1.0147e-05$ ). This result is consistent with experiment I, despite we changed the GAN set to a higher resolution version. Although the same set of CAN images were used in both Experiment I and II, the responses to the human/computer question is much higher in Experiment II (75\% vs 53\%), which confirms our hypothesis about deferring that question. 
The rank order between Abstract Expressionist, CAN, and Art Basel sets, in regards to being created by artist (Q6), is also consistent with Q1 in experiment 1.

\begin{table}
\centering
\footnotesize
\caption{ Means and standard deviations of responses of Experiment II } 
\label{exp2-result}
\begin{tabular}{|c|c|c|c|c|c|c|}
\hline
 & Q1 (std) & Q2 (std) & Q3 (std) & Q4 (std) & Q5 (std) & Q6 (std) \\ 
Image set & Likeness & Novelty & Surprising & Ambiguity & Complexity & human/computer \\ \hline 
DCGAN~\cite{DCGAN16} (256x256)	& 3.23	 (0.53)& 3.08	(0.50)& 3.21 (0.59) &	3.37   (0.48) &	3.18 (0.63) & 	0.65 (0.17) \\ \hline
CAN &	3.30 (0.43)&	3.27	(0.44)&3.13	(0.46)&3.54	(0.45)&3.34	(0.50)&0.75 (0.14) \\ \hline			Abstract Expresionist 	& 3.38	 (0.43)&  3.03 (0.38)& 	2.95 (0.50)& 	3.17 (0.35)& 	2.90 (0.35)& 	0.85 (0.11)\\ \hline	
Art Basel 2016 &	2.95	 (0.70)& 2.69	 (0.59)& 2.36	 (0.66)& 2.79	 (0.59)& 2.46	 (0.68)& 0.48 (0.23) \\ \hline
\end{tabular}						
\end{table}

\subsection*{Experiment III:} 
This goal of this experiment is to judge aspects related to whether the images generated by CAN can be considered art. We compared the CAN set with both the Art Expressionist and the Art Basel Sets.
This experiment is similar to an experiment conducted by Snapper {\it et al}~\cite{snapper2015your} to determine to what degree human subjects find the works of art to be intentional, having visual structure, communicative, and inspirational. In this experiment an image is shown and the subject is asked the following four questions: 
\begin{description}
\item[Q1:] As I interact with this painting, I start to see the artist's™ intentionality: it looks like it was composed very intentionally. 

\item[Q2:]  As I interact with this painting, I start to see a structure emerging.

\item[Q3:]  Communication: As I interact with this painting, I feel that it is communicating with me.

\item[Q4:]  Inspiration: As I interact with this painting, I feel inspired and elevated.
\end{description}
For each of the question the users answered in a scale from 1 (Strongly Disagree) to 5 (Strongly Agree). The users were asked to look at each image at least 5 second before answering. 21 users participated in this experiment with responses from 10 distinct users for each image. 

\begin{table}
\centering
\caption{ Means and standard deviations of the responses of Experiment III } 
\label{exp3-result}
\begin{tabular}{|c|c|c|c|c|}
\hline
             & Q1 (std) & Q2 (std)  & Q3 (std) & Q4 (std) \\
Painting set & Intentionality & Visual Structure & Communication & Inspiration \\ \hline
CAN                      & 3.3 (0.47) & 3.2  (0.47) & 2.7 (0.46) & 2.5 (0.41)  \\\hline
Abstract Expressionist   & 2.8 (0.43)& 2.6 (0.35) & 2.4	(0.41)& 2.3 (0.27)    \\\hline
Art Basel 2016               & 2.5 (0.72)& 2.4 (0.64) & 2.1	(0.59)& 1.9(0.54)      \\\hline
Artist sets combined    & 2.7 (0.6)& 2.5 (0.52)  & 2.2	(0.54)& 2.1 (0.45)     \\\hline
\end{tabular}
\end{table}

Snapper {\it et al}  hypothesized that subjects would rate works by real artists higher in these scales that works by children or animals, and indeed their experiments validated their hypothesis\cite{snapper2015your}. 

We also hypothesized that human subjects would rate art by real artists higher on these scales than those generated by the proposed system. To our surprise the results showed that our hypothesis is not true! Human subjects rated the images generated by the proposed system higher than those created by real artists, whether in the Abstract Expressionism set or in the Art Basel set (see Table~\ref{exp2-result}). 

\begin{figure*}[tph]
\footnotesize
\center
\includegraphics[width=0.9\textwidth]{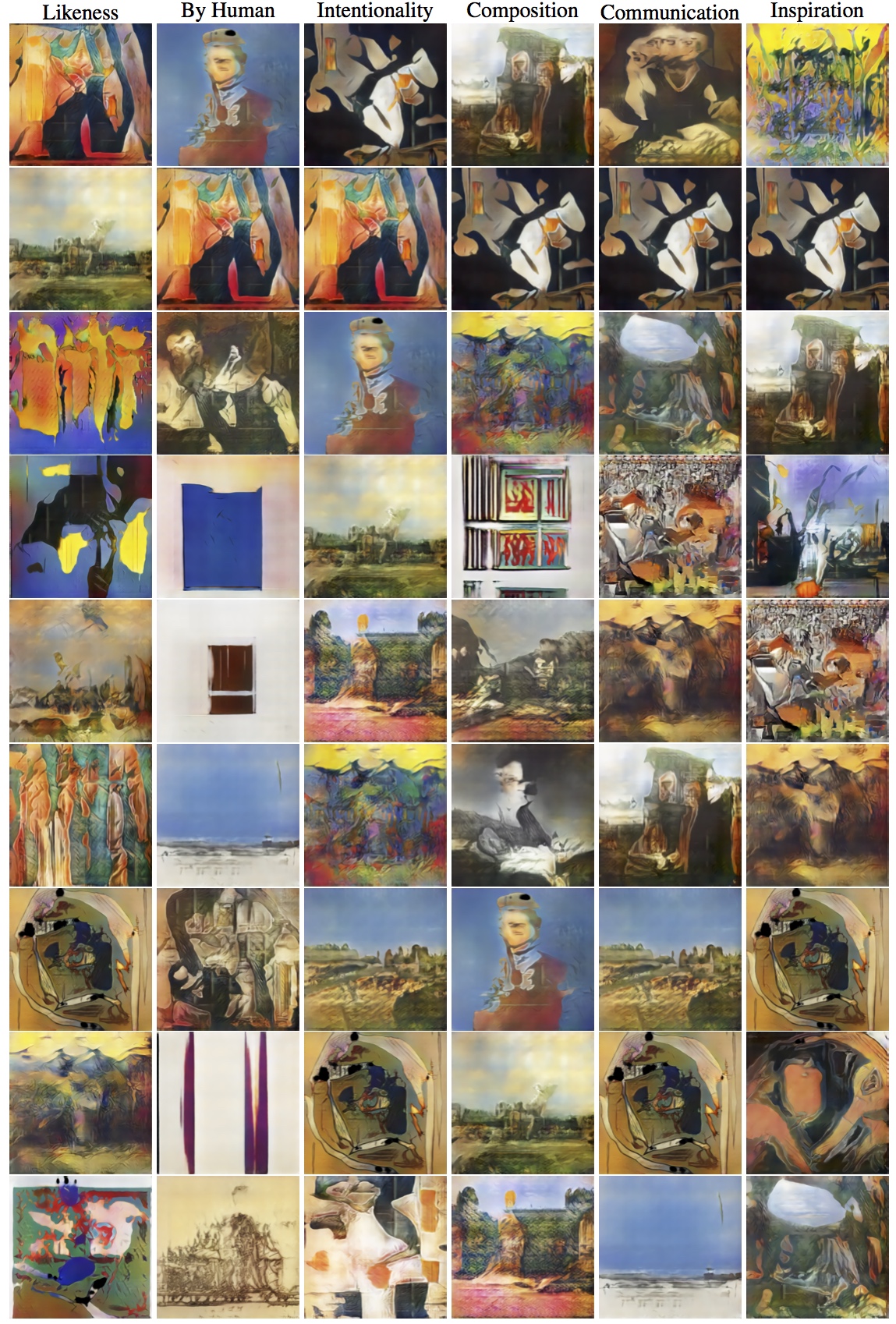} 
\caption{ Top Ranked Images From CAN in Human Subject Experiment I and III}
\label{fig:CANSamples}
\end{figure*}

It might be debatable what a higher score in each of these scales actually means, and whether the differences are statistically significant. However, the fact that subjects found the images generated by the machine intentional, visually structured, communicative, and inspiring, with similar levels to actual human art, indicates that subjects see these images as art! Figure~\ref{fig:CANSamples} shows several examples generated by CAN, ranked by the responses of human subjects to each question.

\subsection*{Experiment IV:}  The goal of this experiment is to evaluate the effect of adding the style ambiguity loss to the CAN model, in contrast to the style classification loss, in generating novel and aesthetically appealing images. In other words, is it learning about styles or deviating from style that causes the results to be creative. To assess creativity we refer to the most common definition of creativity of an artifact as being novel and influential~\cite{PaulBarry2014Ch1,Elgammal2015}. However, since influence is not relevant here, we use novelty as a proxy for creativity. In this experiment, in order to evaluate novelty we used a pool of art history students as sophisticated art-educated subjects who can judge the novelty and aesthetics better than general MTurk subjects. Each subject was shown pairs of images, one from the CAN set and one from the {\it sc}-CAN model, randomly selected and placed in random order side by side. Each subject was asked two questions for each pair:
\begin{description}
\item [Q1] Which image do you think is more novel? 
\item [Q2] Which image do you think is more aesthetically appealing? 
\end{description}
Results: The results of this experiment shows that 59.47\% of the time subjects selected CAN images as more novel and 60\% of the time they found CAN images more aesthetically appealing. 
 This indicates the effect of the style ambiguity loss in the process of generation by CAN compared to the style classification loss.

%

\section{Discussion and Conclusion}

We proposed a system for generating art with creative characteristics. We demonstrated a realization of this system based on a novel creative adversarial network. The system is trained using a large collection of art images from the 15th century to 21st century with their style labels.  The system is able to generate art by optimizing a criterion that maximizes stylistic ambiguity while staying within the art distribution. The system was evaluated by human subject experiments which showed that human subjects regularly confused the generated art with the human art, and sometimes rated the generated art  higher on various high-level scales.

What creative characteristics does the proposed system have? Colton 2008 suggested three criteria that a creative system should have: the ability to produce novel artifacts (imagination), the ability to generate quality artifacts (skill), and the ability to assess its own creation~\cite{colton2008creativity}. Our proposed system possesses the ability to produce novel artifacts because the interaction between the two signals that derive the generation process is designed to force the system to explore creative space to find solutions that deviate from established styles but stay close enough to the boundary of art to be recognized as art. This interaction also provides a way for the system to self-assess its products. The quality of the artifacts is verified by the human subject experiments, which showed that subjects not only thought these artifacts were created by artists, but also rated them higher on some scales than human art.

One of the main characteristics of the proposed system is that it learns about the history of art in its process to create art. However it does not have any semantic understanding of art behind the concept of styles. It does not know anything about subject matter, or explicit models of elements or principle of art. The learning here is based only on exposure to art and concepts of styles. In that sense the system has the ability to continuously learn from new art and would then be able to adapt its generation based on what it learns.

We leave open how to interpret the human subjects' responses that ranked the CAN art better than the Art Basel samples in different aspects. Is it because the users have typical style-backward bias?  Are the subjects biased by their aesthetic assessment? Would that mean that the results are not that creative? More experiments are definitely needed to help answer these questions.







\bibliographystyle{plain}

\bibliography{iccc}

\end{document}